\documentclass[10pt,twocolumn,letterpaper]{article}

\usepackage[pagenumbers]{wacv} %

\usepackage{graphicx}
\usepackage{amsmath}
\usepackage{amssymb}
\usepackage{booktabs}
\usepackage{relsize}
\usepackage{bbold}

\usepackage[pagebackref,breaklinks,colorlinks]{hyperref}

\usepackage[linesnumbered,ruled,vlined]{algorithm2e}
\usepackage{multirow}
\usepackage{colortbl}
\usepackage[most]{tcolorbox}
\usepackage{xcolor}
\newtcbox{\inlinebox}[1][]{
    on line,
    colback=gray!25,
    colframe=gray!25, %
    boxrule=0pt, %
    arc=2pt, %
    boxsep=2pt, %
    left=0pt, right=0pt, top=0pt, bottom=0pt, %
    #1
}
\SetKwInput{KwMethods}{Poisoning methods}

\usepackage[capitalize]{cleveref}
\crefname{section}{Sec.}{Secs.}
\Crefname{section}{Section}{Sections}
\Crefname{table}{Table}{Tables}
\crefname{table}{Tab.}{Tabs.}

\def\wacvPaperID{1644} %
\def\confName{WACV}
\def\confYear{2025}

\begin{document}

\title{LogicNet: A Logical Consistency Embedded Face Attribute Learning Network}

\author{Haiyu Wu$^{1}$, Sicong Tian$^{2}$, Huayu Li$^3$, Kevin W. Bowyer$^{1}$\\
$^{1}$University of Notre Dame\\ $^2$Indiana University South Bend\\ $^3$University of Arizona\\}

\maketitle

\begin{abstract}
Ensuring logical consistency in predictions is a crucial yet overlooked aspect in face attribute classification. We explore the potential reasons for this oversight and introduce two pressing challenges to the field: 1) How can we ensure that a model, when trained with data checked for logical consistency, yields predictions that are logically consistent? 2) How can we achieve the same with training data that hasn't undergone logical consistency checks? Minimizing manual effort is also essential for enhancing automation. To address these challenges, we introduce two datasets, FH41K and CelebA-logic, and propose LogicNet, which combines adversarial learning and label poisoning to learn the logical relationship between attributes without need for post-processing steps.
Accuracy of LogicNet surpasses that of the next-best approach by 13.36\%, 9.96\%, and 1.01\% on FH37K, FH41K, and CelebA-logic, respectively. 
In real-world case analysis, our approach can achieve a reduction of more than 50\% in the average number of failed cases (logically inconsistent attributes) compared to other methods. 
Code link: \url{https://github.com/HaiyuWu/LogicNet/tree/main}.
\vspace{-3mm}
\end{abstract}
    
\section{Introduction}
\label{sec:intro}

\textbf{\textit{Where does the logical consistency problem arise in computer vision?}}
In the realm of multi-attribute classification, models are trained to predict the attributes represented in a given image. Examples include facial attributes~\cite{celeba, pubfig, logical-consistency-fh}, clothing styles~\cite{deepfashion, DCSA}, autonomous driving~\cite{traffic-sign}, human action recognition~\cite{human-action}, and others.
Whenever multiple attributes are predicted for an image, logical relationships may potentially exist among these attributes.
For instance, in a popular scheme for predicting attributes in face images \cite{celeba}, the attributes {\small\tt goatee}, {\small\tt no beard}, and {\small\tt mustache} are predicted independently.
Note that, in CelebA, mustache is a type of beard based on the ground truth.
Logically, then, if {\small\tt no beard} is predicted as true, then both {\small\tt goatee} and {\small\tt mustache} should be predicted as false. 
Logical consistency issues may arise in more subtle interactions as well.
For example, if {\small\tt wearing hat} is predicted as true, then the information required to make a prediction for {\small\tt bald} is occluded.
Similarly, if {\small\tt wearing sunglasses} is predicted as true, then the information to predict {\small\tt narrow eyes} or {\small\tt eyes closed} is occluded.
Analogous logical relationships exist in other datasets.
In DeepFashion~\cite{deepfashion}, it would be logically inconsistent to predict both {\small\tt long-sleeve} and {\small\tt sleeveless}  as true for the same garment. 
In ROAD-R~\cite{traffic-sign}, {\small\tt red light} and {\small\tt green light} are mutually exclusive.
\textit{Issues of logical consistency can arise in various areas of computer vision, particularly in the area of multi-attribute classification}.

\begin{figure}[t]
    \centering
        \includegraphics[width=0.7\linewidth]{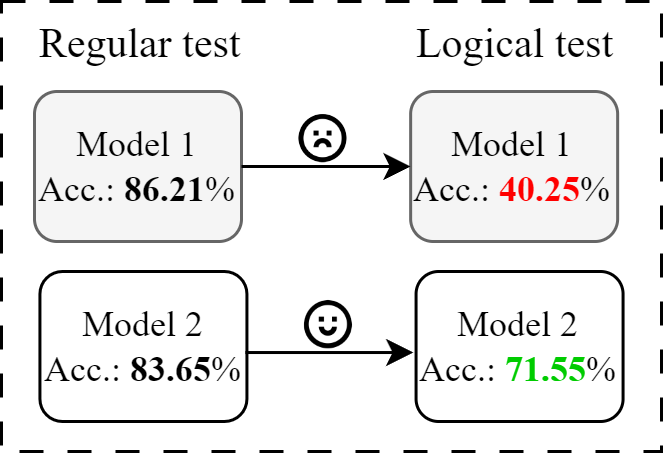}
    \vspace{-3mm}
   \caption{Regular and logical tests on two model types. Model 1 and Model 2 trained without and with logical relationships between predicted results, respectively. In real-world deployment, Model 2 provides more reliable predictions.}
\vspace{-6mm}
\label{fig:teaser}
\end{figure}

\textbf{\textit{Why is logical consistency of predictions important in face attribute classification?}}
Face and body attributes have been extensively utilized in various research domains, including face matching/recognition~\cite{attribute-fr1,attribute-fr2,attribute-fr3,attribute-fr4,pubfig,attribute-fr6}, re-identification~\cite{attribute-re-id1,attribute-re-id2,attribute-re-id3}, training GANs~\cite{attribute-gan1, attribute-gan2, attribute-gan3, attribute-gan4}, bias analysis~\cite{logical-consistency-fh, bias-attribute1, attribute-in-bias, beard-in-bias, bias-attribute2}, and others.
For a fair accuracy comparison across demographic groups, it is important to balance the distribution of non-demographic attributes across the groups \cite{bias-attribute2}.
For example, if a group wants to understand how facial hair affects face recognition accuracy across demographic groups, they have to tightly control variation on facial hair. However, if a model predicts \{{\small\tt clean-shaven} and {\small\tt beard-length-short}\} or \{{\small\tt beard-at-chin-area} and {\small\tt full-beard}\} for the same image, this type of prediction will put same images in two conflicting categories and significantly impact the validity of results. 
To train a face editing GAN, face attributes are used to guide the model to generate the target images~\cite{attgan, stgan, attr-ddpm}.
However, if images exhibit logically inconsistent sets of attribute values, these applications of the attributes become problematic and prone to errors.
Hence, 
\textit{logical consistency of attribute  predictions is crucial for essentially all higher-level computer vision tasks}. 

\textit{\textbf{Why has logical consistency not received more attention in face attribute classification?}} 
1) \textit{Attribute design}: Unlike traffic signs and objects, the attributes in the existing face attribute datasets are more ambiguous~\cite{celeba-consistency}, such as high cheekbones, attractive, oval face, which makes it hard to determine the logical relationship between attributes. 2) \textit{Accuracy measurement metric}: Accuracy is commonly measured by $\frac{N_{tp} + N_{tn}}{N_{total}}$, which considers predictions independently across attributes. In multi-attribute classification, this measurement ignores the logical relationships and hides the imbalanced performance on positive and negative samples. Hence, the logical consistency problem cannot be reflected by accuracy numbers. 3) \textit{Ground truth accuracy}: Wu et al.~\cite{celeba-consistency} and Lingenfelter et al.~\cite{celeba-consistency-emily} reported on accuracy problems with CelebA labels. Since the model performance is only measured in the aforementioned way, involving logical relationships might not boost, or could even hurt, the accuracy.
These three reasons could give rise to the situation where none of the recent face attribute survey papers~\cite{facial-attr-survey1, facial-attr-survey2, facial-attr-survey3, facial-attr-survey4} even mentions the problem of logical consistency.

This paper introduces two challenging tasks for face attribute classification:
(1) Training a model with \textbf{labels in the training data that have been checked for logical consistency} (FH37K and FH41K), aiming to improve the accuracy and logical consistency of predictions; and (2) Training a model with \textbf{labels in the training data that have not been checked for logical consistency} (CelebA-logic). No post-processing steps should be used in these tasks.
Contributions of this work include:
\vspace{-1mm}
\begin{itemize}
    \item \textit{A set of logical-relationship-cleaned annotations for CelebA validation and testing sets to support a more challenging task: train a logically consistent model with logical-consistency-unchecked data.}
    \vspace{-1mm}
    \item \textit{A larger benchmark dataset, FH41K, which compensates the lack of samples of the minority group in the existing facial hair dataset.}
    \vspace{-1mm}
    \item \textit{Proposing a new training method, LogicNet, which combining adversarial learning and data poisoning algorithms to achieve higher accuracy and lower logical inconsistency across three face attribute datasets without using any post-processing steps.}
\end{itemize}

\section{Related Work}
\paragraph{\texttt{Method comparison}}
Fischer et al.~\cite{dl2} defined a superclass which represents a group of classes in order to make the misclassifications less severe. However, it aims to solve the problem in multi-class classification rather than multi-label classification.
Xu et al.~\cite{semantic-loss} utilized the confidence values of the predictions and logical relationships to model the semantic loss, and added it as a regularization term to the classification loss. 
Giunchiglia et al.~\cite{traffic-sign} proposed a constraint loss and a post-processing step, constraint output, to make the model predict safer/more logical.
Wu et al.~\cite{logical-consistency-fh} proposed a Logical Consistency Prediction loss (LCPloss) to add more punishments to the logically inconsistent predictions and a post-processing step to correct some of the logically inconsistent predictions. \textit{These methods make the model "afraid" to predict positive labels to achieve a high accuracy}, due to the sparsity of the dataset labels and most of the logical relationships occurring between positive labels. This makes the model less useful in the real-world and needs human to design rules for prediction correction. Different from them, our method generates rich logically inconsistent labels for training, so the model learns logical relationships while predicting more positive labels. Moreover, no additional post-processing is needed.
\vspace{-4mm}
\paragraph{\texttt{Face attribute datasets:}} Kumar et al.~\cite{pubfig} added 73 face attributes to the LFW~\cite{lfw} dataset. They also collected 58K images with each image assigned 73 attributes to form the PubFig~\cite{pubfig}. The labels in these two datasets are not checked for logical consistency and the dataset scale is small. CelebA~\cite{celeba} is the most prevalent face attribute dataset with 200K images, where each image has 40 attributes. The other works after it use the same set of attributes, for example, the University of Maryland Attribute Evaluation Dataset (UMD-AED)~\cite{umd-aed}, Youtube Faces Dataset\cite{youtube_face_dataset}, MAAD~\cite{maad}, and CelebV-HQ~\cite{celebv}. However, none of them check the logical consistency of the labels.
Wu et al.~\cite{logical-consistency-fh} proposed the FH37K dataset, which has 37K images with 22 attributes for each image. Importantly, the labels are checked for logical consistency, so it is used as a benchmark in this paper.

\section{Benchmarks}
\label{sec:benchmarks}

The ambiguity of attribute names and the reasons listed in Section~\ref{sec:intro} limit the research in evaluating model performance on the dimension of logical consistency. We enlarge the FH37K dataset and clean the test and validation sets of CelebA for future benchmark tests. The rules are listed in the Algo. 1 and 2 of Supplementary Material.

\noindent
\textbf{FH37K}~\cite{logical-consistency-fh} is the first dataset checking both logical consistency and accuracy of the annotations. It contains 37,565 images, coming from a subset of CelebA~\cite{celeba} and a subset of WebFace260M~\cite{webface260m}. Each image has 22 attributes of facial hair and baldness. However, due to the small number of positive samples of attributes ``Bald Sides Only'' (161 images), ``Long'' beard (747 images), and ``Bald Top and Sides'' (1,321 images), insufficient train/val/test samples limits the performance measurement on logical consistency.
To address this, we augment this dataset by adding more positive samples to minority classes.
\begin{figure}[t]
    \centering
        \includegraphics[width=\linewidth]{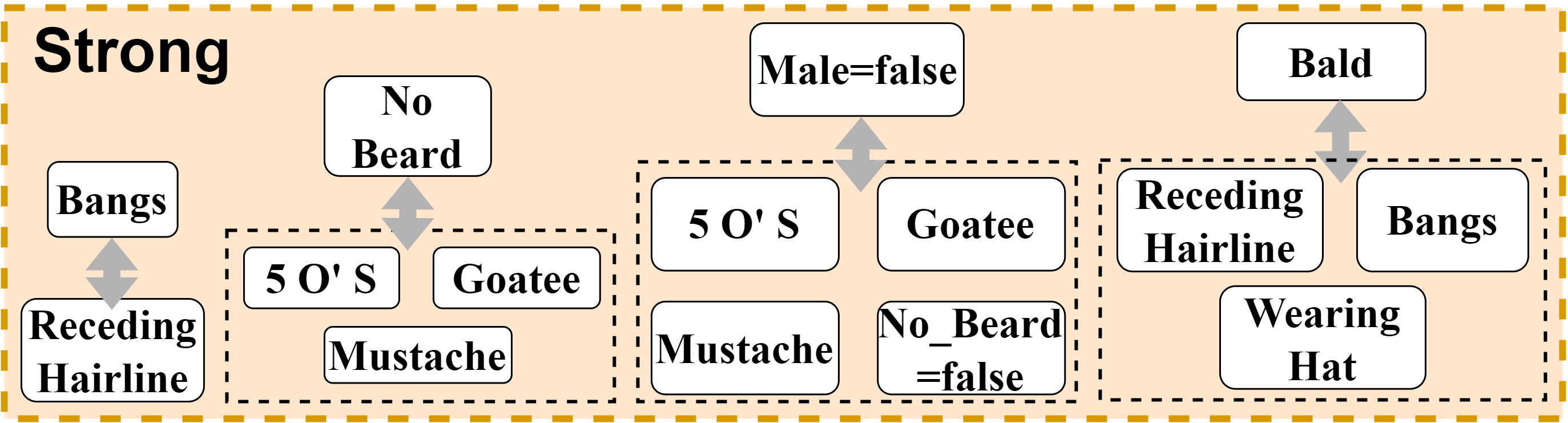}
    \vspace{-5mm}
   \caption{Strong logical relationship between attributes in CelebA. The attributes are split to three categories. \textbf{Strong}: Impossible in most cases. \textbf{Weak}: Rarely possible in some cases. \textbf{Independent/Ambiguous}: The attributes are either ambiguous on definitions~\cite{celeba-consistency, celeba-consistency-emily}. The latter two are in the Supplementary Material.}
\label{fig:logical-relationships}
\vspace{-4mm}
\end{figure}

\noindent
\textbf{FH41K} is our extended version of FH37K.
We added 3,712 images from 2,096 identities from WebFace260M to increase the number of positive examples of attributes that had too low of a representation in FH37K. Specifically, we used the best facial hair classification model trained with FH37K\footnote{https://github.com/HaiyuWu/LogicalConsistency\#testing}
to select the images that have confidence value higher than 0.8 for both "Long" and for "Bald Sides Only". We then engaged a human annotator, with the prior knowledge learnt from the documentation provided by ~\cite{logical-consistency-fh}, to manually check the selected images in order to ensure the accuracy and logical consistency of the added images. Major changes are in Table~\ref{tab:sample-changes}.

Both FH37K and FH41K have a set of rigorously defined rules based on the logical relationships including \textit{mutually exclusive}, \textit{dependency}, and \textit{collectively exhaustive}. The annotations are evaluated based on these relationships. 
However, previous existing face attribute datasets were created without considering the issue of logical consistency in annotations. For example, in CelebA, there are 1,503 images having ``Bald=true" and ``Receding\_Hairline=true'', 235 images having both "Mustache=true" and "No\_Beard=true", and 139 images having ``Male=false" and ``No\_Beard=false".
This raises an important question.  
\textbf{\textit{Is it possible to train a model to produce logically consistent attribute predictions using a training dataset that does not have logically consistent annotations?}} 
We compiled an additional dataset based on CelebA to study this question.

\noindent
\textbf{CelebA-logic} is the variation of CelebA, where the logical relationship between attributes is checked for both validation and test sets, as shown in Table~\ref{tab:celeba-changes}. Given the absence of a definitive guide of how these 40 attributes are marked and what the definition is for each attribute, we categorized the attribute relationships into three groups, based on our knowledge, as shown in Figure~\ref{fig:logical-relationships}. To make a fair set of logical rules, only \textbf{Strong} relationships are used to check the logical consistency. Moreover, ~\cite{celeba-consistency-emily, celeba-consistency} reported that CelebA suffers from a substantial rate of inaccurate annotations. Hence, we conducted an annotation cleaning process for those strong relationship attributes upon the MSO-cleaned-annotations~\cite{celeba-consistency}. To get the cleaned facial hair and baldness related attributes, we converted the FH37K annotations back to the CelebA version and updated the labels to the corresponding images. Two human annotators then marked ``Bangs'', ``Receding Hairline'' and ``Male'' based on the designed definitions for all the images in the validation and test sets. To ensure the consistency and accuracy of the new annotations, a third human annotator with knowledge of definitions  marked 1,000 randomly selected samples. The estimated consistency rate is 88.73\%, measured by Eq.~\ref{eq:acc}. Consequently, 1) all images are cropped and aligned to 224x224 based on the given landmarks,
2) 975 images are omitted from the original dataset, 3) \textbf{63,557 (31.8\%)} images have at least one different label than the original, and 4) all test and validation annotations obey the \textbf{Strong} logical relationships.
\setlength{\tabcolsep}{0.5mm}

\begin{figure}[t]
\centering
    \begin{minipage}{0.5\textwidth}
        \centering
        \begin{tabular}{c|c|c}
        \hline
        \multirow{2}{*}{Attributes}                    & \multicolumn{2}{c}{\# positive samples}  \\ \cline{2-3}
         & FH37K & FH41K \\ \hline
         Bald Sides Only & 161 & 1,316\\
         Long & 747 & 2,509\\
         Bald Top and Sides & 1,321 & 1,880\\ \hline
        \end{tabular}
        \vspace{-3mm}
        \captionof{table}{The number of positive samples of under-represented attributes between FH37K and FH41K.}
        \label{tab:sample-changes}
        \vspace{1mm}
    \end{minipage}
    \begin{minipage}{0.5\textwidth}
        \centering
        \begin{tabular}{c|c|c|c}
        \hline
         Datasets                    & Train & Val & Test  \\ \hline
         CelebA & - & - & -\\
         CelebA-logic & - & cleaned & cleaned\\ \hline
        \end{tabular}
        \vspace{-3mm}
        \captionof{table}{Changes between CelebA and CelebA-logic. ``-" means logical consistency is not manually checked.}
        \label{tab:celeba-changes}
    \end{minipage}
    \vspace{-6mm}
\end{figure}

\vspace{-3mm}

\section{Proposed Method}
To provide a  solution for the challenges of logical consistency, we propose LogicNet, which exploits an adversarial training strategy and a label poisoning algorithm, Bag of Labels (BoL).
LogicNet enables the classifier to learn the logical relationships between attributes, thereby enhancing the model's capacity to generate logically consistent predictions.
\begin{figure}[t]
    \centering
        \includegraphics[width=\linewidth]{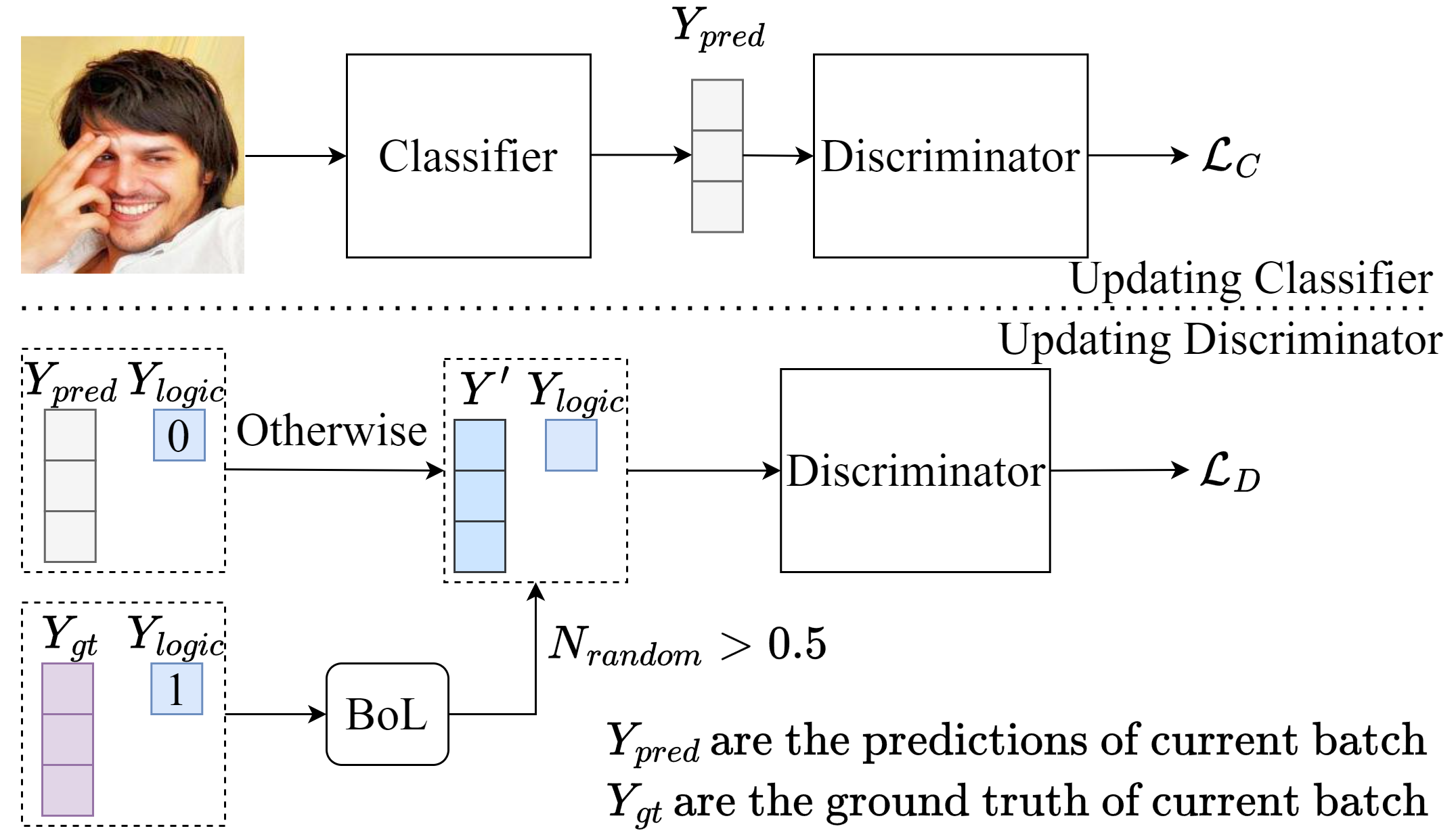}
    \vspace{-3mm}
   \caption{The proposed LogicNet. The weights of multi-attribute classifier and discriminator are updated alternatively. $Y'$ is either the predictions of the classifier or the poisoned labels from BoL algorithm. $Y_{logic}$ is the logical consistency label vector.}
\label{fig:algo}
\vspace{-3mm}
\end{figure}

\vspace{-2mm}
\subsection{Adversarial Training}

We propose an adversarial training framework, shown in Figure~\ref{fig:algo}, to compel the classifier $\mathcal{C}$ to make logically consistent predictions while improving the accuracy of predictions. 
Formalizing the desired goal,
we consider a set of training images as $X\in \{x_1, x_2,..., x_N\}$, 
from which we want to train a model, 
$\mathcal{F}(X)$, to project $X$ to the ground truth labels $Y_{gt}\in\{y_1, y_2,...,y_N\}$, where each $y_N$ is a set of attribute labels of $x_N$. The classification loss is the binary cross entropy loss:
\vspace{-4mm}
\begin{equation}
    \begin{split}
    \mathcal{L}_{bce}(\mathcal{F}(X;\Phi), Y_{gt}) = &-\frac{1}{N}\sum^{N}_{i=1}[y_{i}log(\mathcal{F}(x_i;\Phi))\\&+
    (1-y_{i})log(1-\mathcal{F}(x_i;\Phi))]
    \end{split}
\end{equation}
Where $\Phi$ is the parameter vector of the classifier $\mathcal{C}$. For the adversarial learning, a discriminator that can judge the logical consistency of the predictions is needed. Here, we use a multi-headed self-attention network~\cite{transformer} to give a probability, $\mathcal{P}_{logic}\in [0, 1]$, for the logical consistency of labels, $Y'$. The loss of the multi-attribute classifier, $\mathcal{L}_C$, becomes:
\begin{equation}
    \underset{\Phi}{\min} \underset{\Theta}{\max}(1-\lambda)\mathcal{L}_{bce}(\mathcal{F}(X;\Phi), Y_{gt})+\lambda log(-\mathcal{D}(\mathcal{F}(X);\Theta))
\end{equation}
Where $\mathcal{D}$ is the parameter frozen discriminator, $\Theta$ is the parameter vector of the discriminator, and $\lambda$ is used for loss trade-off.

\subsection{Bag of Labels}
To train a discriminator, the straightforward approach~\cite{correlation3} is to directly feed the predictions (ground truth labels) and treat them as negative (positive) samples. Since the training labels of CelebA are not yet cleaned, using them could mislead the discriminator and cause it to learn incorrect patterns. 
Hence, we propose a Bag of Labels algorithm, as shown in Algorithm~\ref{algo:bol}, that can automatically poison the ground truth labels to generate logically {\it inconsistent} labels while detecting the logical consistency of the original label. This algorithm consists with two parts: Condition Group Setup and Label Poisoning.

\begin{algorithm}[t]
\SetKwFunction{FMain}{BagOfLabels}
\SetKwProg{Fn}{Function}{:}{}
\Fn{\FMain{$Y_{gt}$, $Y_{logic}$, $dataset$}}{
    \KwIn{\\ a) $Y_{gt}$: attribute ground truth labels \\ b) $Y_{logic}$: initialized logic labels \\ c) $dataset$: dataset name}
    \KwOut{\\ a) $Y_{bol}$: the generated labels \\ b) $Y_{logic}$: the ground truth labels of logic group}
    \KwMethods{\\ a) $inter\_imp\_poison$: generating impossible inter-class labels\\ b) $intra\_imp\_poison$: generating impossible intra-class labels\\ c) $intra\_incomp\_poison$: generating incomplete intra-class labels }
    \DontPrintSemicolon
    Initialize condition groups: $g_{c1}$ and $g_{c2}$ \;
    \If{$dataset != celeba$}{
        Randomly split $Y_{gt}$ to $Y^{1}_{gt}$, $Y^{2}_{gt2}$, $Y^{3}_{gt3}$ \;
        $Y_{bol}^{1}=inter\_imp\_poison(Y^{1}_{gt}, g_{c1}, g_{c2})$ \;
        $Y_{bol}^{2}=intra\_imp\_poison(Y^{2}_{gt})$ \;
        $Y_{bol}^{3}=intra\_incomp\_poison(Y^{3}_{gt})$ \;
        Update $Y_{logic}$ in each method. \;
        Merge $Y_{bol}^{1}$, $Y_{bol}^{2}$, $Y_{bol}^{3}$ to form $Y_{bol}$
    }
    \Else{
        Randomly split $Y_{gt}$ to $Y^{1}_{gt}$, $Y^{2}_{gt}$,..., $Y^{len(g_{c1})}_{gt}$ \;
        \For{$i$ in $range(len(g_{c1}))$}{
            $Y_{bol}=inter\_imp\_poison(Y^{i}_{gt}, g_{c1}[i], g_{c2}[i])$
        }
        Update $Y_{logic}$ during the loop: $\mathbb{1}_{!poisoned}$
    }
    return $Y_{bol}$, $Y_{logic}$
}
\caption{Bag of Labels}
\label{algo:bol}
\end{algorithm}

\noindent
\textbf{Condition Group Setup:}
To give accurate logic labels $Y_{logic}$ to $Y'$, following the relationships, we separate the corresponding attributes of each rule to two groups: $g_{c1}$ and $g_{c2}$, where the attributes in $g_{c1}[i]$ have strong logical relationships with the attributes in $g_{c2}[i]$. For both FH37K and FH41K, we follow the rules given by~\cite{logical-consistency-fh}. For CelebA, we follow the relationships in Figure~\ref{fig:logical-relationships} and the propositional logic version is in the Algo. 2 of Supplementary Material.

\noindent
\textbf{Label Poisoning:}
To generate logically inconsistent labels, we first categorize the rules in three cases: inter-class impossible poisoning, intra-class impossible poisoning, and intra-class incomplete poisoning. Inter-class impossible poisoning aims to generate labels where the logical inconsistency happens between the attributes in different classes (e.g. Beard Area(clean shaven)=true and Beard Length(short)=true; Bald=true and Receding\_Hairline=true). Intra-class impossible and intra-class incomplete poisoning aim to generate labels where there are multiple positive predictions within one class (e.g. Beard Area(clean shaven)=true and Beard Area(chin area)=true) or no positive predictions within a class. These two poisoning strategies apply to FH37K and FH41K; attributes in CelebA do not have this level of detail and so do not have these logical relationships.
After each poisoning, the initialized logic labels, $Y_{logic}$, are updated on-the-fly. Note that, each poisoning method randomly attacks one relationship at one time. Thus, if the positive attributes are not in the target relationship, no poisoning will happen and $Y_{logic}=1$.

The objective function for the discriminator training is:
\begin{equation}
    \underset{\Theta}{\min}\mathcal{L}_{\mathcal{D}} = \mathcal{L}_{bce}(Y_{logic}, \mathcal{D}(Y', Y_{gt}))
\end{equation}
Where 
\begin{equation}
    Y' =
    \left\{\begin{matrix}
    N_{random} > 0.5, & Y_{bol}\\ 
    Others, & Y_{pred}
    \end{matrix}\right.
\end{equation}
Here, $Y_{bol}$ is from BoL algorithm, $Y_{pred}$ is from classifier, $N_{random}$ is a randomly generated float number between 0 and 1. When using $Y_{bol}$ the target $Y_{logic}$ is from BoL algorithm, otherwise $Y_{logic}=0$.

\section{Experiments}
In this section, we evaluate the proposed approach from two aspects: accuracy and logical consistency. For accuracy, the traditional average accuracy measurement (Eq.~\ref{eq:traditional}, where $N$ = total number of images, $N_{tp}$ = number of true positive predictions, $N_{tn}$ = number of true negative predictions) ignores the imbalanced number of positive and negative images for each attribute. 
\begin{equation}
    AccT_{avg} = \frac{1}{N}(N_{tp} + N_{tn})\times100
         \label{eq:traditional}
\end{equation}
Ignoring the imbalance of positive and negative attributes results in an unfair measure of model performances since the multi-attribute classification datasets suffer from sparse annotations. For example, in the original CelebA annotations, if all predictions are negative, the overall test accuracy is 76.87\%. Hence, we follow the suggestion in~\cite{Imbalanced} and use average value of the positive accuracy, $Acc^{p}_{avg}$, and negative accuracy, $Acc^{n}_{avg}$, to consider the imbalance:
\begin{equation}
    Acc_{avg} = \frac{1}{2}(Acc^{p}_{avg}+Acc^{n}_{avg})
\label{eq:acc}
\end{equation}
In addition, to show how logical consistency of predictions affects the accuracy, we measure the performance under two conditions: 1) \textit{\textbf{without considering the logical consistency on predictions}}, 2) \textit{\textbf{considering logical consistency by counting logically inconsistent predictions as  incorrect}}. For FH37K and FH41K, we also include the label compensation strategy~\cite{logical-consistency-fh} experiments to complete the accuracy comparison. To evaluate the model performance on logical consistency, we performed logical consistency checking on the predictions of 600K images from WebFace260M~\cite{webface260m}. We also independently compare the accuracy values on the strong relationship attributes in CelebA-logic.

To give a comprehensive study of the impact of logical consistency on model predictions, we choose six training methods for the experiment. Binary Cross Entropy Loss (BCE) is a baseline which only considers the entropy between predictions and ground truth labels. Binary Focal Loss (BF)~\cite{focal} aims to focus more on hard samples in order to mitigate the effect of imbalanced data. BCE-MOON~\cite{moon} tries to balance the effect of positive and negative samples by
calculating the ratio of positive and negative samples for each attribute as the weights added to loss values before back propagation. Semantic~\cite{semantic-loss} uses the confidence values of predictions and logical relationships to form a regularization term, in order to constrain the classification loss. Constrained~\cite{traffic-sign} uses the fuzzy logic relaxation of the constraints to constrain the classification loss. Logically Consistent Prediction Loss (LCP)~\cite{logical-consistency-fh} utilizes the conditional probability of the binary predictions to force the probability of the mutually exclusive attributes happening at the same time to be 0 and the probability of the dependency attributes happening at the same time to be 1.

\subsection{Implementation}
We train all the classifiers starting with the pretrained ResNet50~\cite{resnet} from Pytorch\footnote{https://pytorch.org/hub/pytorch\_vision\_resnet/}. The FH37K results in Table~\ref{tab:fh-acc} are adopted from~\cite{logical-consistency-fh} except the values of $Acc_{avg}$. We resize images to 224x224 for all three datasets. The batch size and learning rate are \{256, 0.0001\} for FH37K and FH41K, and \{64, 0.001\} for CelebA-logic. We use random horizontal flip for both FH37K and FH41K. We use random horizontal flip, color jitter, and random rotation for CelebA-logic. AFFACT~\cite{affact} and ALM~\cite{alm} are the two SOTA models that are available online, which we used for performance comparison on CelebA-logic. The $\lambda$ values for FH37K, FH41K, and CelebA are $\{0.15, 0.2, 0.1\}$. The discriminator consists of 8 multi-headed self-attention blocks. 
It is necessary to know that the ALM algorithm resizes the original (178x218) CelebA images to 128x128 for testing; the other methods use the cropped images mentioned in Section~\ref{sec:benchmarks} for testing.

\setlength{\tabcolsep}{0.5mm}
\begin{table*}[t]
    \centering
    \begin{tabular}{l||c|c|c||c|c|c}
        \hline
        \multirow{2}{*}{Methods} & \multicolumn{3}{c||}{FH37K} & \multicolumn{3}{c}{FH41K}\\ \cline{2-7}
         & $Acc_{avg}$ & $Acc^{n}_{avg}$ & $Acc^{p}_{avg}$ & $Acc_{avg}$ & $Acc^{n}_{avg}$ & $Acc^{p}_{avg}$\\ \hline \hline
        \multicolumn{7}{l}{Logical consistency is not taken into account.}\\ \hline
        BCE & 79.23 & 94.72 & 63.73 & 83.88 & \underline{95.50} & 72.27\\ 
        BCE-MOON~\cite{moon} & \textcolor[HTML]{1A09F3}{\textbf{86.21}} & 90.67 & \textcolor[HTML]{1A09F3}{\textbf{81.75}} & \textcolor[HTML]{1A09F3}{\textbf{88.02}} & 91.29 & \textcolor[HTML]{1A09F3}{\textbf{84.75}}\\ 
        BF~\cite{focal} & 76.92 & \underline{95.43} & \textcolor[HTML]{E92341}{58.41} & 75.81 & \textcolor[HTML]{1A09F3}{\textbf{97.78}} & \textcolor[HTML]{E92341}{52.85}\\ 
        \cellcolor{gray!25}Semantic~\cite{semantic-loss} & \cellcolor{gray!25}80.30 & \cellcolor{gray!25}94.26 & \cellcolor{gray!25}66.33 & \cellcolor{gray!25}83.73 & \cellcolor{gray!25}95.52 & \cellcolor{gray!25}71.95\\ 
        \cellcolor{gray!25}Constrained~\cite{traffic-sign} & \cellcolor{gray!25}80.45 & \cellcolor{gray!25}94.20 & \cellcolor{gray!25}66.69 & \cellcolor{gray!25}83.78 & \cellcolor{gray!25}95.43 & \cellcolor{gray!25}72.13\\ 
        \cellcolor{gray!25}LCP~\cite{logical-consistency-fh} & \cellcolor{gray!25}79.64 & \cellcolor{gray!25}\textcolor[HTML]{1A09F3}{\textbf{95.98}} & \cellcolor{gray!25}63.30 & \cellcolor{gray!25}84.93 & \cellcolor{gray!25}95.09 & \cellcolor{gray!25}74.77\\ 
        \cellcolor{gray!25}\textbf{Ours} & \cellcolor{gray!25}\underline{83.65} & \cellcolor{gray!25}93.46 & \cellcolor{gray!25}\underline{73.83} & \cellcolor{gray!25}\underline{85.66} & \cellcolor{gray!25}94.23 & \cellcolor{gray!25}\underline{77.10}\\ \hline \hline
        \multicolumn{7}{l}{With label compensation.}\\ \hline
        BCE$^\dagger$ & \underline{80.14} & \underline{91.49} & 68.78 & 79.12 & 87.43 & 70.81\\ 
        BCE-MOON$^\dagger$ & \textcolor[HTML]{E92341}{42.59} & \textcolor[HTML]{E92341}{50.55} & \textcolor[HTML]{E92341}{34.62} & \textcolor[HTML]{E92341}{42.79} & \textcolor[HTML]{E92341}{47.96} & \textcolor[HTML]{E92341}{37.61}\\ 
        BF$^\dagger$ & 78.48 & 90.91 & 66.05 &  \textcolor[HTML]{1A09F3}{\textbf{82.85}} & \textcolor[HTML]{1A09F3}{\textbf{93.53}} & \underline{73.17}\\ 
        \cellcolor{gray!25}Semantic$^\dagger$ & \cellcolor{gray!25}76.43 & \cellcolor{gray!25}87.50 & \cellcolor{gray!25}65.35 & \cellcolor{gray!25}79.57 & \cellcolor{gray!25}88.07 & \cellcolor{gray!25}71.06\\ 
        \cellcolor{gray!25}Constrained$^\dagger$ & \cellcolor{gray!25}76.28 & \cellcolor{gray!25}87.23 & \cellcolor{gray!25}65.33 & \cellcolor{gray!25}79.17 & \cellcolor{gray!25}87.67 & \cellcolor{gray!25}70.68 \\ 
        \cellcolor{gray!25}LCP$^\dagger$ & \cellcolor{gray!25}\textcolor[HTML]{1A09F3}{\textbf{81.44}} & \cellcolor{gray!25}\textcolor[HTML]{1A09F3}{\textbf{92.65}} & \cellcolor{gray!25}\textcolor[HTML]{1A09F3}{\textbf{70.23}} & \cellcolor{gray!25}79.31 & \cellcolor{gray!25}87.31 & \cellcolor{gray!25}71.31\\ 
        \cellcolor{gray!25}\textbf{Ours$^\dagger$} & \cellcolor{gray!25}78.28 & \cellcolor{gray!25}87.23 & \cellcolor{gray!25}\underline{69.32} & \cellcolor{gray!25}\underline{81.53} & \cellcolor{gray!25}\underline{89.10} & \cellcolor{gray!25}\textcolor[HTML]{1A09F3}{\textbf{73.96}}\\ \hline \hline
        \multicolumn{7}{l}{Without label compensation. (\textcolor[HTML]{FF4081}{\textbf{{A general solution}}})}\\ \hline 
        BCE$^\dagger$ & \textcolor[HTML]{E92341}{48.50} & \textcolor[HTML]{E92341}{54.59} & \textcolor[HTML]{E92341}{42.40} & \textcolor[HTML]{E92341}{56.71} & 62.14 & \textcolor[HTML]{E92341}{51.27}\\ 
        BCE-MOON$^\dagger$ & \textcolor[HTML]{E92341}{40.25} & \textcolor[HTML]{E92341}{47.54} & \textcolor[HTML]{E92341}{32.95} & \textcolor[HTML]{E92341}{40.68} & \textcolor[HTML]{E92341}{45.39} & \textcolor[HTML]{E92341}{35.98}\\ 
        BF$^\dagger$ & \textcolor[HTML]{E92341}{36.20} & \textcolor[HTML]{E92341}{40.95} & \textcolor[HTML]{E92341}{31.45} & \textcolor[HTML]{E92341}{22.38} & \textcolor[HTML]{E92341}{23.84} & \textcolor[HTML]{E92341}{20.92}\\ 
        \cellcolor{gray!25}Semantic$^\dagger$ & \cellcolor{gray!25}\textcolor[HTML]{E92341}{57.74} & \cellcolor{gray!25}65.87 & \cellcolor{gray!25}\textcolor[HTML]{E92341}{49.62} & \cellcolor{gray!25}\textcolor[HTML]{E92341}{56.73} & \cellcolor{gray!25}62.36 & \cellcolor{gray!25}\textcolor[HTML]{E92341}{51.10}\\ 
        \cellcolor{gray!25}Constrained$^\dagger$ & \cellcolor{gray!25}\textcolor[HTML]{E92341}{58.19} & \cellcolor{gray!25}66.14 & \cellcolor{gray!25}\textcolor[HTML]{E92341}{50.25} & \cellcolor{gray!25}\textcolor[HTML]{E92341}{57.25} & \cellcolor{gray!25}62.94 & \cellcolor{gray!25}\textcolor[HTML]{E92341}{51.56}\\ 
        \cellcolor{gray!25}LCP$^\dagger$ & \cellcolor{gray!25}\textcolor[HTML]{E92341}{38.69} & \cellcolor{gray!25}\textcolor[HTML]{E92341}{43.70} & \cellcolor{gray!25}\textcolor[HTML]{E92341}{33.67} & \cellcolor{gray!25}\underline{64.54} & \cellcolor{gray!25}\underline{70.40} & \cellcolor{gray!25}\textcolor[HTML]{E92341}{58.67}\\ 
       \cellcolor{gray!25}\textbf{Ours$^\dagger$} & \cellcolor{gray!25}\textcolor[HTML]{1A09F3}{\textbf{71.55}} & \cellcolor{gray!25}\textcolor[HTML]{1A09F3}{\textbf{79.37}} & \cellcolor{gray!25}\textcolor[HTML]{1A09F3}{\textbf{63.73}} & \cellcolor{gray!25}\textcolor[HTML]{1A09F3}{\textbf{74.50}} & \cellcolor{gray!25}\textcolor[HTML]{1A09F3}{\textbf{81.41}} & \cellcolor{gray!25}\textcolor[HTML]{1A09F3}{\textbf{67.59}}\\ \hline
    \end{tabular}
    \vspace{-3mm}
    \caption{Accuracy of models trained with different methods on FH37K (left) and FH41K (right) dataset. $\dagger$ means the measurements consider the logical consistency. [Keys: \textcolor[HTML]{1A09F3}{\textbf{Best}}, \textcolor[HTML]{E92341}{$<$60\%}, \underline{Second best}, \inlinebox{Logic involved methods}].
    }
    \label{tab:fh-acc}
\end{table*}

\setlength{\tabcolsep}{0.5mm}
\begin{table*}[t]
    \centering
    \begin{tabular}{c|c|c|c||c|c|c}
    \hline
        \multirow{2}*{Methods} &  \multicolumn{3}{c||}{W/o considering logical consistency} & \multicolumn{3}{|c}{Considering logical consistency} \\ \cline{2-7}
         & $Acc_{avg}$ & $Acc^{n}_{avg} $ & $Acc^{p}_{avg}$ & $Acc_{avg}$ & $Acc^{n}_{avg} $ & $Acc^{p}_{avg}$ \\ \hline
        AFFACT (original) & 81.25 & 95.72 & 66.78 & 79.11 & 93.55 & 64.67\\ 
        ALM (original) & 81.97 & 94.25 & 69.69 & 79.04 & 91.04 & 67.03\\ \hline \hline
        AFFACT & 79.71 & \underline{95.48} & 63.95 & 77.72 & \underline{93.31} & 62.12\\ 
        ALM & 80.53 & 94.10 & 66.95 & 77.63 & 90.88 & 64.39\\ 
        BCE &  80.89 & 94.96 & 66.70 & 77.94 & 92.34 & 63.54\\ 
        BCE-MOON & \textcolor[HTML]{1A09F3}{\textbf{87.13}} & 87.95 & \textcolor[HTML]{1A09F3}{\textbf{86.32}} & 74.76 & 76.24 & \textcolor[HTML]{1A09F3}{\textbf{73.28}}\\ 
        BF & 76.44 & \textcolor[HTML]{1A09F3}{\textbf{96.77}} & 56.11 & 75.28 & \textcolor[HTML]{1A09F3}{\textbf{95.82}} & 54.75\\ 
        \cellcolor{gray!25}Semantic & \cellcolor{gray!25}80.66 & \cellcolor{gray!25}94.61 & \cellcolor{gray!25}66.71 & \cellcolor{gray!25}77.72 & \cellcolor{gray!25}91.94 & \cellcolor{gray!25}63.50\\ 
        \cellcolor{gray!25}Constrained & \cellcolor{gray!25}80.54 & \cellcolor{gray!25}94.91 & \cellcolor{gray!25}66.18 & \cellcolor{gray!25}77.81 & \cellcolor{gray!25}92.35 & \cellcolor{gray!25}63.27\\ 
        \cellcolor{gray!25}LCP & \cellcolor{gray!25}81.91 & \cellcolor{gray!25}94.16 & \cellcolor{gray!25}69.66 & \cellcolor{gray!25}\underline{78.07} & \cellcolor{gray!25}90.26 & \cellcolor{gray!25}65.87\\ 
        \cellcolor{gray!25}\textbf{Ours} & \cellcolor{gray!25}\underline{82.18} & \cellcolor{gray!25}93.74  & \cellcolor{gray!25}\underline{70.63} & \cellcolor{gray!25}\textcolor[HTML]{1A09F3}{\textbf{79.08}} & \cellcolor{gray!25}90.89 & \cellcolor{gray!25}\underline{67.28}\\ \hline
    \end{tabular}
    \vspace{-3mm}
    \caption{Accuracy of models trained with different methods on CelebA-logic dataset. "original" means the model is tested with the same images but using the original annotations. [Keys: \textcolor[HTML]{1A09F3}{\textbf{Best}}, \underline{Second best}, \inlinebox{Logic involved methods}]}
    \vspace{-4mm}
    \label{tab:celeba-logic-acc}
\end{table*}

\subsection{Accuracy}
\setlength{\tabcolsep}{0.5mm}
\begin{table*}[t]
    \centering
    \begin{tabular}{c|c|c|c|c|c|c|c|c|c|c}
    \hline
        Methods & 5 O' S & Bald & Bangs & Goatee & Male & Mustache & No\_Beard & $^*$Hairline & $^*$Hat & $Acc_{avg}$\\ \hline
        AFFACT & \underline{72.24} & \textcolor[HTML]{1A09F3}{\textbf{90.27}} & 85.36 & 70.41 & \underline{95.51} & 61.86 & \underline{85.47} & \underline{65.69} & \textcolor[HTML]{1A09F3}{\textbf{93.93}} & 80.08\\
        ALM & \textcolor[HTML]{1A09F3}{\textbf{76.34}} & 81.81 & 85.08 & 74.27 & 93.88 & 64.51 & \textcolor[HTML]{1A09F3}{\textbf{87.66}} & 62.84 & 90.62 & 79.67\\ \hline \hline
        BCE & 65.88 & 75.68 & \underline{86.69} & 76.73 & 94.78 & 87.80 & 80.68 & \textcolor[HTML]{1A09F3}{\textbf{67.28}} & 89.96 & 80.61\\
        BCE-MOON & 69.79 & 70.77 & 82.22 & \textcolor[HTML]{1A09F3}{\textbf{80.73}} & 82.09 & 82.73 & 77.40 & 59.67 & 84.20 & 76.62\\
        BF & 59.38 & 78.05 & 78.52 & 69.33 & \textcolor[HTML]{1A09F3}{\textbf{96.82}} & 87.59  & 83.12 & 62.88 & \underline{92.13} & 78.65\\
        \cellcolor{gray!25}Semantic & \cellcolor{gray!25}65.34 & \cellcolor{gray!25}77.78 & \cellcolor{gray!25}86.87 & \cellcolor{gray!25}76.95 & \cellcolor{gray!25}94.36 & \cellcolor{gray!25}88.21 & \cellcolor{gray!25}80.54 & \cellcolor{gray!25}64.98 & \cellcolor{gray!25}89.96 & \cellcolor{gray!25}80.55\\
        \cellcolor{gray!25}Constrained & \cellcolor{gray!25}66.61 & \cellcolor{gray!25}78.82 & \cellcolor{gray!25}85.73 & \cellcolor{gray!25}71.73 & \cellcolor{gray!25}94.76 & \cellcolor{gray!25}87.76 & \cellcolor{gray!25}80.42 & \cellcolor{gray!25}65.50 & \cellcolor{gray!25}91.06 & \cellcolor{gray!25}80.26\\
        \cellcolor{gray!25}LCP & \cellcolor{gray!25}69.83 & \cellcolor{gray!25}82.91 & \cellcolor{gray!25}84.00 & \cellcolor{gray!25}75.33 & \cellcolor{gray!25}92.72 & \cellcolor{gray!25}\underline{88.54} & \cellcolor{gray!25}81.30 & \cellcolor{gray!25}63.57 & \cellcolor{gray!25}89.70 & \cellcolor{gray!25}\underline{80.88}\\
        \cellcolor{gray!25}\textbf{Ours} & \cellcolor{gray!25}68.10 & \cellcolor{gray!25}\underline{86.03} & \cellcolor{gray!25}\textcolor[HTML]{1A09F3}{\textbf{87.79}} & \cellcolor{gray!25}\underline{79.05} & \cellcolor{gray!25}94.54 & \cellcolor{gray!25}\textcolor[HTML]{1A09F3}{\textbf{89.40}} & \cellcolor{gray!25}81.45 & \cellcolor{gray!25}65.55 & \cellcolor{gray!25}91.39 & \cellcolor{gray!25}\textcolor[HTML]{1A09F3}{\textbf{82.59}}\\ \hline
    \end{tabular}
    \vspace{-3mm}
    \caption{Accuracy values of the attributes that have strong logical relationships in CelebA-logic with considering the logical consistency. 5 O' S, $^*$Hairline, and $^*$Hat are 5 O'clock Shadow, Receding Hairline and Wearing Hat. [Keys: \textcolor[HTML]{1A09F3}{\textbf{Best}}, \underline{Second best}, \inlinebox{Logic involved methods}]}
    \label{tab:strong-relation-acc}
\end{table*}

\setlength{\tabcolsep}{0.5mm}
\begin{table*}[h]
    \centering
    \begin{tabular}{l|c|c|c||c|c|c}
    \hline
         \multicolumn{1}{c|}{\multirow{2}{*}{Methods}} &  \multicolumn{3}{c||}{W/o considering logical consistency} & \multicolumn{3}{|c}{Considering logical consistency} \\ \cline{2-7}
         & $Acc_{avg}$ & $Acc^{n}_{avg} $ & $Acc^{p}_{avg}$ & $Acc_{avg}$ & $Acc^{n}_{avg} $ & $Acc^{p}_{avg}$ \\ \hline
        Adversarial learning (preds) & 82.63 & 93.42 & 71.83 & 65.94 & 74.18 & 57.70\\ 
        Reward model (BoL) & 81.90 & 93.04 & 70.77 & 65.04 & 73.23 & 56.86\\ 
        LogicNet (preds + BoL) & \textcolor[HTML]{1A09F3}{\textbf{83.65}} & \textcolor[HTML]{1A09F3}{\textbf{93.46}} & \textcolor[HTML]{1A09F3}{\textbf{73.83}} & \textcolor[HTML]{1A09F3}{\textbf{71.55}} & \textcolor[HTML]{1A09F3}{\textbf{79.37}} & \textcolor[HTML]{1A09F3}{\textbf{63.73}}\\ \hline \hline
        Adversarial learning  (preds) & \textcolor[HTML]{1A09F3}{\textbf{85.72}} & 94.12 & \textcolor[HTML]{1A09F3}{\textbf{77.32}} & 72.83 & 79.38 & 66.28\\ 
        Reward model (BoL) & 85.48 & 94.05 & 76.91 & 73.03 & 79.96 & 66.11\\ 
        LogicNet (preds + BoL) & 85.66 & \textcolor[HTML]{1A09F3}{\textbf{94.23}} & 77.10 & \textcolor[HTML]{1A09F3}{\textbf{74.50}} & \textcolor[HTML]{1A09F3}{\textbf{81.41}} & \textcolor[HTML]{1A09F3}{\textbf{67.59}}\\ \hline \hline
        Adversarial learning  (preds) & 81.46 & \textcolor[HTML]{1A09F3}{\textbf{94.42}} & 68.49 & 77.95 & 91.07 & 64.82\\ 
        Reward model (BoL) & 80.65 & 94.18 & 67.12 & 78.15 & \textcolor[HTML]{1A09F3}{\textbf{91.72}} & 64.58\\ 
        LogicNet (preds + BoL) & \textcolor[HTML]{1A09F3}{\textbf{82.18}} & 93.74 & \textcolor[HTML]{1A09F3}{\textbf{70.63}} & \textcolor[HTML]{1A09F3}{\textbf{79.08}} & 90.89 & \textcolor[HTML]{1A09F3}{\textbf{67.28}}\\ \hline
    \end{tabular}
    \vspace{-3mm}
    \caption{Ablation study for training a logic discriminator resulting in the accuracy of the classifier. The testing sets are FH37K (Top), FH41K (Middle), CelebA-logic (Bottom). [Keys: \textcolor[HTML]{1A09F3}{\textbf{Best}}]}
    \vspace{-4mm}
    \label{tab:ablation}
\end{table*}

Table~\ref{tab:fh-acc} and Table~\ref{tab:celeba-logic-acc} show the accuracy values, tested on FH37K, FH41K, and CelebA-logic, under two measurement conditions. In the traditional case of not considering logical consistency of predictions, every method reaches $>75\%$ average accuracy, 
However, when logically inconsistent predictions are counted as incorrect, 
 accuracy decreases  across all training methods.

For FH37K and FH41K, the average decreases in accuracy of logical-inconsistency-ignored methods (i.e., BCE, BCE-MOON, and BF) are 39.14\% and 42.65\%, and the average decreases of logical-consistency-aware methods (i.e. Semantic, Constrained, and LCP) are 28.59\% and 23.06\%.  The proposed method has \{12.1\%, 11.16\%\} decrease in accuracy and the overall accuracy is \{13.36\%, 9.96\%\} higher than the second-highest accuracy, and \{35.35\%, 52.12\%\} higher than the lowest accuracy. The results show that, although incorporating logical relationship in the objective function improves the model's ability to make logically consistent predictions, logical consistency of predictions is still a serious problem. The proposed method reduces the disparity better than the existing methods.

~\cite{logical-consistency-fh} proposed a post-processing step, termed label compensation strategy, to resolve  incomplete predictions. By using this strategy, the methods other than BCE-MOON have a significant increase in accuracy, \{26.70\% and 24.91\%\} on average. This results in two conclusions: 1) Methods that aim to mitigate the imbalanced data effect might give an illusion of high accuracy driven by positive predictions; 2) Other methods can somewhat catch the logical patterns, but need to involve post-processing steps. However, the label compensation strategy is only for solving the collectively exhaustive case (i.e., the model must give one positive prediction in an attribute group). For example, in FH37K and FH41K, the attributes, \{clean-shaven, chin-area, side-to-side, beard-area-information-not-visible\}, in the Beard Area group can cover any case that is related to beard area. Implementing this type of strategy necessitates extensive manual analysis to determine the most judicious decision-making process, underscoring the imperative for continued research in this domain.

For CelebA-logic, when considering logical consistency of predictions, the patterns echo the previous observations. For both AFFACT and ALM, we use the original model weights provided by the authors. The top half of Table~\ref{tab:celeba-logic-acc} shows that either using the original or the cleaned test annotations, there is a 2.49\% accuracy decrease after considering logical consistency. \textbf{Note that the overall increase in accuracy since CelebA was published is 5.98\%,  so requiring logical consistency erases about half of the improvement seen on CelebA since it was introduced.
}
The average accuracy decrease of the models tested on the cleaned annotations is 4.04\%, where the BF has the smallest accuracy difference and BCE-MOON has the largest accuracy difference. Our speculation is that BF over-focuses on negative attributes but the logical relationship mostly happens between positive side, so BF has lower probability to violate logical relationships. Conversely, BCE-MOON over-focuses on positive side, so it has higher probability to disobey logical relationships.
Results in Table~\ref{tab:celeba-logic-acc} and Table~\ref{tab:strong-relation-acc} show that the proposed method has the best performances on average accuracy of all attributes and strong relationship attributes, where it is \{1.01\%, 1.71\%\} higher than the second-highest accuracy. Therefore, the proposed method has the best ability to obtain the logical relationships.

The results could be summarized in the following points: 1) Without any post-processing steps, our method performs the best, on average, than either \textbf{general objective functions} (i.e., BCE, MOON, BF) or \textbf{logical relationship embedded objective functions} (i.e., Semantic, Constrained, LCP) across three datasets. 2) Adversarial training implemented in our method does not bring large computing overhead. 3) 1.01\% improvement on CelebA-logic dataset is not trivial, as there are 40 attributes and the overall improvement on the original CelebA dataset since it was published is around 5\%. In addition, the evaluation of logical consistency on predictions in real-world application and comparison with GPT4-V are in the Supplementary Material.
\begin{table}[t]
    \centering
    \begin{tabular}{c|c|c|c|c|c}
        \hline
         & baseline & Semantic & Constrained & LCP & Ours\\ \hline
        $t_{train}$ & 1.78h & 8.12h & 9.77h & 5.32h & 3.6h\\ \hline
    \end{tabular}
    \vspace{-3mm}
    \caption{Training time for 20 epochs on CelebA-logic. Hardware: 4 RTX6000 GPUs. Backbone: ResNet50. Baseline: BCE method.
    }
    \label{table:running-time}
\end{table}

\subsection{Discussion}
\paragraph{\textbf{Is LogicNet proper?}} Simply using adversarial learning does not achieve the best performance because the classifier eventually learns good patterns. Always regarding predictions as logically inconsistent can lead to model collapse. Purely using the poisoned labels from the BoL algorithm to train the discriminator makes the discriminator work as a reward model. It helps the classifier learn logical relationships but can easily reach the local minimum. LogicNet combines these two, which surpass the performance of implementing each individually, as shown in Table~\ref{tab:ablation}. Table~\ref{table:running-time} shows that LogicNet brings the least computational complexity compared to other methods. Hence, it is more efficient and effective than other methods.

\paragraph{\textbf{Is LogicNet necessary, if the post-processing works?}} Although LogicNet does not perform the best after involving post-processing steps, the complexity of pre-training design for each method is different. For any of the ``logic-involved methods", they need to list 19 circumstances for FH37K and FH41K training and 20 circumstances for CelebA training, whereas LogicNet only needs 4 conditions for the label poisoning of each dataset training. Note that as the number of attributes increases, the complexity increases rapidly. Therefore, LogicNet can significantly reduce the complexity of pre-training design.

\paragraph{\textbf{Why do we choose CelebA and FH37K?}} First, FH37K is the only dataset checking the logical consistency of the ground truth labels, so it fits our task well. Second, there are larger datasets, \emph{e.g.}, MAAD and CelebV-HQ, but they use the same set of CelebA attributes and the logical consistency of labels is not checked. To provide a dataset for the second of the aforementioned challenge, the validation and testing sets need to be cleaned. To this end, CelebA is a proper dataset with sufficient images and relatively easy to manually correct the validation and testing data.

\paragraph{\textbf{Why don't we compare with other SOTA methods?}} There is significant number of works trying to improve the performance on CelebA dataset~\cite{position-relationship1, position-relationship2,correlation1, correlation2, correlation3,logical1}. However, unlike these works doing the regular test on CelebA, which does not need the weights from other works, the logical test in this paper needs to retrain the model and test them on the cleaned test sets. Unfortunately, \textbf{none of them} provide official implementation/weights for their methods. Under this situation, we tried our best to ask for weights and re-implement methods, so AFFACT~\cite{affact}, ALM~\cite{alm}, MOON~\cite{moon}, Semantic loss~\cite{semantic-loss} and Constrained loss~\cite{traffic-sign} can be compared in this paper. It is impossible for us to re-implement all other methods.

\paragraph{\textbf{Limitations}} LogicNet provides a general solution to cause model predictions to be more logically consistent than previous methods, but the accuracy difference before and after considering logical consistency of predictions is still large and the failed ratio is not negligible for both challenges. Moreover, LogicNet needs more epochs to be well-trained than other methods. Specifically, the validation accuracy of other methods mostly saturates around 10 epochs, but LogicNet needs around 18 epochs to get the best validation accuracy.

\section{Conclusions and Limitations}

We provide two new datasets for two logical consistency challenges: 1) Train a classifier with logical-consistency-checked data to learn a classifier that makes logically consistent predictions, and 2) Train a classifier with training data that contains logically inconsistent labels and still achieves logically consistent predictions.
To our best knowledge, this is the first work that comprehensively discusses the problem of logical consistency of predictions in face attribute classification.

We propose the LogicNet, which does not involve any post-processing steps and significantly improves performance, \{13.36\% (FH37K), 9.96\% (FH41K), 1.01\% (CelebA-logic), 1.71\% (CelebA-logic 9 attributes)\} compared to the second best, under logical-consistency-checked condition for all three datasets. For the real-world case analysis, the proposed method can substantially reduce the fraction of failed (logically inconsistent) predictions.

{\small
\bibliographystyle{ieee_fullname}
\bibliography{main}
}

\end{document}


\title{\LaTeX\ Author Guidelines for \confName~Proceedings}

\author{First Author\\
Institution1\\
Institution1 address\\
{\tt\small firstauthor@i1.org}
\and
Second Author\\
Institution2\\
First line of institution2 address\\
{\tt\small secondauthor@i2.org}
}
\maketitle

\section{Logical Consistency Evaluation in Real World}

\begin{table*}[t]
    \centering
    \begin{tabular}{l||c|c|c||c|c|c}
        \hline
        \multirow{2}{*}{Methods} & \multicolumn{3}{c||}{FH37K} & \multicolumn{3}{c}{FH41K}\\ \cline{2-7}
         & $N_{incomp}$ & $N_{imp}$ & $R_{failed}$ & $N_{incomp}$ & $N_{imp}$ & $R_{failed}$\\ \hline \hline
        \multicolumn{7}{l}{With label compensation.}\\ \hline
        BCE & 0 & 11,134 & 1.84 & 0 & 7,464 & 1.24\\ 
        BCE-MOON & 0 & 330,115 & \textcolor[HTML]{E92341}{54.66} & 0 & 341,114 & \textcolor[HTML]{E92341}{56.48}\\ 
        BF & 0 & 14,007 & 2.32 & 0 & 3,530 & \textcolor[HTML]{1A09F3}{\textbf{0.58}}\\ 
        \cellcolor{gray!25}Semantic & \cellcolor{gray!25}0 & \cellcolor{gray!25}36,372 & \cellcolor{gray!25}6.02 & \cellcolor{gray!25}0 & \cellcolor{gray!25}44,803 & \cellcolor{gray!25}7.42\\ 
        \cellcolor{gray!25}Constrained & \cellcolor{gray!25}0 & \cellcolor{gray!25}38,971 & \cellcolor{gray!25}6.45 & \cellcolor{gray!25}0 & \cellcolor{gray!25}46,601 & \cellcolor{gray!25}7.72\\ 
        \cellcolor{gray!25}LCP & \cellcolor{gray!25}0 & \cellcolor{gray!25}5,595 & \cellcolor{gray!25}\textcolor[HTML]{1A09F3}{\textbf{0.93}} & \cellcolor{gray!25}0 & \cellcolor{gray!25}5,788 & \cellcolor{gray!25}0.96\\ 
        \cellcolor{gray!25}\textbf{Ours} & \cellcolor{gray!25}0 & \cellcolor{gray!25}21,731 & \cellcolor{gray!25}3.60 & \cellcolor{gray!25}0 & \cellcolor{gray!25}19,194 & \cellcolor{gray!25}3.18\\ \hline \hline
        \multicolumn{7}{l}{Without label compensation. (\textcolor[HTML]{FF4081}{\textbf{A general solution}})}\\ \hline
        BCE & 240,761 & 6,001 & 40.86 & 352,061 & 585 & \textcolor[HTML]{E92341}{58.39}\\ 
        BCE-MOON & 31,512 & 313,044 & \textcolor[HTML]{E92341}{57.05} & 34,415 & 321,872 & \textcolor[HTML]{E92341}{59.00}\\ 
        BF & 339,136 & 1,295 & \textcolor[HTML]{E92341}{56.37} & 587,056 & 0 & \textcolor[HTML]{E92341}{97.21}\\ 
        \cellcolor{gray!25}Semantic & \cellcolor{gray!25}185,824 & \cellcolor{gray!25}21,040 & \cellcolor{gray!25}34.25 & \cellcolor{gray!25}227,558 & \cellcolor{gray!25}21,482 & \cellcolor{gray!25}41.24\\ 
        \cellcolor{gray!25}Constrained & \cellcolor{gray!25}171,319 & \cellcolor{gray!25}23,255 & \cellcolor{gray!25}32.22 & \cellcolor{gray!25}203,063 & \cellcolor{gray!25}26,390 & \cellcolor{gray!25}37.99\\ 
        \cellcolor{gray!25}LCP & \cellcolor{gray!25}307,576 & \cellcolor{gray!25}300 & \cellcolor{gray!25}\textcolor[HTML]{E92341}{50.98} & \cellcolor{gray!25}248,768 & \cellcolor{gray!25}2,416 & \cellcolor{gray!25}41.59\\ 
        \cellcolor{gray!25}\textbf{Ours} & \cellcolor{gray!25}139,184 & \cellcolor{gray!25}14,660 & \cellcolor{gray!25}\textcolor[HTML]{1A09F3}{\textbf{25.47}} & \cellcolor{gray!25}133,245 & \cellcolor{gray!25}13,838 & \cellcolor{gray!25}\textcolor[HTML]{1A09F3}{\textbf{24.36}}\\ \hline
    \end{tabular}
    \caption{Logical consistency test on predictions. The models are trained with FH37K (left) and FH41K (right). $N_{incomp}$, $N_{imp}$, and $R_{failed}$ are the number of incomplete predictions, the number of impossible predictions, and failed ratio. [Keys: \textcolor[HTML]{1A09F3}{\textbf{Best}}, \textcolor[HTML]{E92341}{\textbf{$>50\%$}}, \inlinebox{Logic involved methods}]}
    \vspace{-4mm}
    \label{tab:fh-consistency}
\end{table*}

\begin{table*}[h]
    \centering
    \begin{tabular}{c|c|c|c|c}
    \hline
        \multirow{2}*{Methods} & \multicolumn{4}{|c}{Considering logical consistency} \\ \cline{2-5}
         & $N_{imp}$ & $R_{failed}$ & $N^{p}_{mean}$ & $N^{p}_{std.}$\\ \hline
        AFFACT & 5,038 & 0.83 & 1.65 & $\pm$0.68\\ \hline
        ALM & 85 & \textcolor[HTML]{1A09F3}{\textbf{0.01}} & 1.58 & $\pm$0.5\\ \hline \hline
        BCE & 8,723 & 1.44 & 1.53 & $\pm$0.65 \\ \hline
        BCE-MOON & 167,261 & \textcolor[HTML]{E92341}{\textbf{27.70}} & 2.41 & $\pm$1.09 \\ \hline
        BF & 6,897 & 1.14 & 1.22 & $\pm$0.45 \\ \hline
        \cellcolor{gray!25}Semantic & \cellcolor{gray!25}4,187 & \cellcolor{gray!25}0.69 & \cellcolor{gray!25}1.62 & \cellcolor{gray!25}$\pm$0.64 \\ \hline
        \cellcolor{gray!25}Constrained & \cellcolor{gray!25}4,703 & \cellcolor{gray!25}0.78 & \cellcolor{gray!25}1.76 & \cellcolor{gray!25}$\pm$0.68 \\ \hline
        \cellcolor{gray!25}BCE+LCP & \cellcolor{gray!25}3,008 & \cellcolor{gray!25}0.5 & \cellcolor{gray!25}1.81 & \cellcolor{gray!25}$\pm$0.62 \\ \hline
        \cellcolor{gray!25}Ours & \cellcolor{gray!25}5,541 & \cellcolor{gray!25}0.92 & \cellcolor{gray!25}1.62 & \cellcolor{gray!25}$\pm$0.64 \\ \hline
    \end{tabular}
    \caption{Logical consistency test of the algorithms trained with CelebA-logic. $N_{incomp}$, $R_{failed}$, $N^{p}_{mean}$, $N^{p}_{std.}$ are the number of incomplete predictions, failed ratio, the average and standard deviation number of positive predictions on 9 attributes that have strong logical relationships. [Keys: \textcolor[HTML]{1A09F3}{\textbf{Best}}, \textcolor[HTML]{E92341}{\textbf{Worst}}, \inlinebox{Logic involved methods}]}
    \label{tab:celeba-logic-consistency}
\end{table*}

To evaluate the performance of logical consistency of predictions in the real-world case, we use the subset of WebFace260M, which contains 603,910 images, as a test set. Since there are no ground truth labels, we only measure the ratio of failed (logically inconsistent) predictions for each method. 

Table~\ref{tab:fh-consistency} shows that without the post-processing step, the logical-inconsistency-ignored methods have failed ratios \{51.43\% and 71.53\} on FH37K and FH41K, the logical-consistency-aware methods have the failed ratios \{39.15\%, 40.27\%\}, the proposed method significantly reduces the number of failed cases, with a failed ratio, \{25.47\%, 24.36\%\}, that is less than half of the average failed ratio of no-logic-involved methods and $>13\%$ lower than logic-involved methods. BF trained with FH41K predicts too many negative labels which causes the outlier ratio, 97\%. When we implement the post-processing strategy, all the incomplete cases are gone, which results in a low failed ratio for all methods other than BCE-MOON.  This supports the speculation in the main paper, where BCE-MOON over-focuses on positive side and existing methods can somewhat learn the pattern but need to involve post-processing steps.

Table~\ref{tab:celeba-logic-consistency} shows the logical consistency test results on CelebA attributes. Since incomplete prediction is not the case for CelebA attributes, only the number of impossible predictions is used to measure the failed ratio. The results show that, except BCE-MOON, all the other methods have $<2\%$ failed ratio. To dig out the rationale, we also calculate the mean and standard deviation of the number of positive predictions for the attributes that have strong logical relationships. All the methods, other than BCE-MOON, have less than 2 positive predictions on average. This number of positive predictions has limited probability to cause the disobedience of logical consistency.
Moreover, the images in WebFace260M are 112x112, which is different from the resolution of the training images of any methods. Therefore, this test result is \textbf{not considered} in this paper and \textbf{will not be included} in the future work.

\section{Comparing with GPT4-V}
We also tested the performance of GPT4-V~\cite{gpt4v}, the best Vision-Language model, on these three datasets. Since GPT4-V API has a limitation of monthly spend, all the FH37K and FH41K images and 1,000 randomly sampled CelebA images are tested. Note that the super-resolution option in GPT4-V is not used in this experiment. We tried two types of response formats: 1) returning the binary version attribute predictions of the images, 2) returning the attribute names of the positive predictions. Format 1 works better for FH37K and FH41K and format 2 works better for CelebA. The average accuracy on FH37K, FH41K, and CelebA are \{51.86\%, 51.61\%, 56.84\%\} without checking logical consistency, and \{48.10\%, 49.29\%, 54.05\%\} with checking logical consistency of predictions. 
The prompt we used for getting the predictions is in the Code~\ref{code:celeb} and \ref{code:fh}. 
This low performance could be caused by several factors - 1) Accuracy measurement: reporting the average accuracy of positive and negative predictions evenly reveals the uneven performance. For example, in the traditional way, GPT4-V will has 80.62\% accuracy, but the accuracy on the positive side is only 18.62\%. 2) Attribute ambiguity: there is no documentation guiding GPT4-V to better understand the logical relationships underneath, which makes it hard to make correct/logical predictions. 3) Image resolution: the original images are in 112x112 but GPT4-V needs 512x512 images, so the performance drops significantly. Consequently, the current Vision-Language models cannot handle this challenge well. 

\section{Others}

\begin{figure}[t]
    \centering
        \includegraphics[width=1\linewidth]{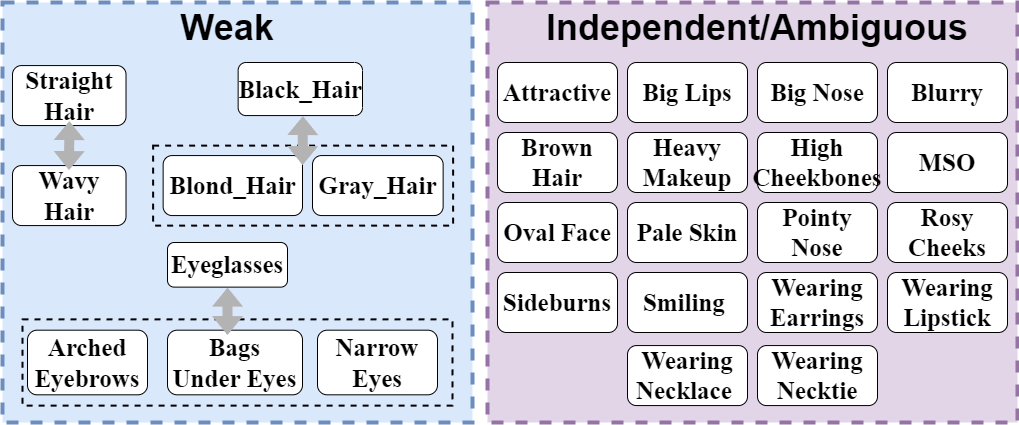}
   \caption{Weak and independent/ambiguous attributes in CelebA.}
\label{fig:ambiguous}
\end{figure}

Algorithm~\ref{algo:fh_fail_detection} and \ref{algo:celeba-fail_detection} are the logical rules used to detect the logically inconsistent predictions. We adopt the rules from~\cite{logical-consistency-fh} for FH37K and FH41K. These rules are used in both accuracy measurement and real-world experiments. Figure~\ref{fig:ambiguous} shows the attributes in the CelebA dataset that have weak logical relationships, and are independent from the other attributes (e.g., Mouth Slightly Open, Wearing Earrings, Wearing Necktie) or have ambiguous definitions (e.g., Attractive, Oval Face, Blurry). 
\begin{algorithm}[t]
    \caption{FH37K/41K Failed prediction detection}
    \label{algo:fh_fail_detection}
    \begin{algorithmic}
        \STATE \textbf{Attribute groups (Category: List$_{\text{attr}}$)}
        \STATE \hspace{\algorithmicindent}\textit{Beard areas}: Clean Shaven, Chin Area, Side to Side, Info not Vis
        \STATE \hspace{\algorithmicindent}\textit{Beard lengths}: 5 O'clock Shadow, Median, Long, Info not Vis
        \STATE \hspace{\algorithmicindent}\textit{Mustache}: None, Isolated, Connected-to-beard, Info not Vis
        \STATE \hspace{\algorithmicindent}\textit{Sideburns}: None, Present, Connected-to-beard, Info not Vis
        \STATE \hspace{\algorithmicindent}\textit{Bald}: False, Top only, Sides only, Top and Sides, Info not Vis
        \STATE \textbf{Fail conditions}
        \STATE \hspace{\algorithmicindent}\textit{Mutually exclusive}:
        \STATE \hspace{\algorithmicindent}\hspace{\algorithmicindent}1. More than one positive prediction in Beard areas (except Info not Vis), 
        \STATE \hspace{\algorithmicindent}\hspace{\algorithmicindent}\hspace{\algorithmicindent}Beard lengths (except Info not Vis), Mustache, Sideburns, Bald group
        \STATE \hspace{\algorithmicindent}\hspace{\algorithmicindent}2. Clean Shaven + any of Beard lengths/Mustache \STATE \hspace{\algorithmicindent}\hspace{\algorithmicindent}\hspace{\algorithmicindent}Connected-to-beard/Sideburns Connected-to-beard
        \STATE \hspace{\algorithmicindent}\hspace{\algorithmicindent}3. Chin area + Sideburns Connected-to-beard
        \STATE \hspace{\algorithmicindent}\hspace{\algorithmicindent}4. Bald (Top and Sides or Sides only) + having sideburns (Sideburns Present, 
        \STATE \hspace{\algorithmicindent}\hspace{\algorithmicindent}\hspace{\algorithmicindent}Sideburns Connected-to-beard)
        \STATE \hspace{\algorithmicindent}\textit{Dependency}:
        \STATE \hspace{\algorithmicindent}\hspace{\algorithmicindent}1. Having beard (Chin Area, Side to Side) + one of the beard lengths must be 
        \STATE \hspace{\algorithmicindent}\hspace{\algorithmicindent}\hspace{\algorithmicindent}true
        \STATE \hspace{\algorithmicindent}\hspace{\algorithmicindent}2. Mustache is connected to beard + $\neg$(Chin Area, Side to Side)
        \STATE \hspace{\algorithmicindent}\hspace{\algorithmicindent}3. Sideburns is connected to beard + $\neg$Side to Side
        \STATE \hspace{\algorithmicindent}\textit{Collectively exhaustive}
        \STATE \hspace{\algorithmicindent}\hspace{\algorithmicindent}No positive prediction in Beard area/Beard lengths/Mustache/Sideburns/Bald
    \end{algorithmic}
\end{algorithm}

\begin{algorithm}
    \caption{CelebA Failed prediction detection}
	\begin{algorithmic}
        \STATE \textbf{Attribute Groups (attr: List$_{attr}$)}
	    \STATE \ \ \ \ \ No Beard: 5 O'clock Shadow, Goatee, Mustache
        \STATE \ \ \ \ \ $\neg$Male: 5 O'clock Shadow, Goatee, Mustache, $\neg$No Beard
        \STATE \ \ \ \ \ Bangs: Receding Hairline
        \STATE \ \ \ \ \ Bald: Receding Hairline, Bangs, Wearing Hat
        \STATE \textbf{Logic rules}
        \STATE \ \ \ \ \ \textit{Mutually exclusive}: $((attr\wedge\neg List_{attr})\vee(\neg attr\wedge List_{attr}))$
	    \STATE \textbf{Fail conditions:} $attr\wedge List_{attr}$
	\end{algorithmic}
    \label{algo:celeba-fail_detection}
\end{algorithm}

\begin{listing}[h]
\relsize{-3}
\centering
\begin{minted}{python}
content = [
    {
        "type": "text",
        "text": f"""Give the predictions of [5_o_Clock_Shadow,Arched_Eyebrows,Attractive,Bags_Under_Eyes,Bald,Bangs,Big_Lips,
            Big_Nose,Black_Hair,Blond_Hair,Blurry,Brown_Hair,Bushy_Eyebrows,Chubby,Double_Chin,Eyeglasses,
            Goatee,Gray_Hair,Heavy_Makeup,High_Cheekbones,Male,Mouth_Slightly_Open,Mustache,
            Narrow_Eyes,No_Beard,Oval_Face,Pale_Skin,Pointy_Nose,Receding_Hairline,Rosy_Cheeks,
            Sideburns,Smiling,Straight_Hair,Wavy_Hair,Wearing_Earrings,Wearing_Hat,Wearing_Lipstick,
            Wearing_Necklace,Wearing_Necktie,Young] for each image. Try your best to predict.
            You should give me the positive predictions all the other attributes will be considered as negative.
            Output format: 'image name - No_Beard,Wavy_Hair,...,Bald' for each image.
            Self-check: 
            1) make sure you consider the logical relationship.
            Try you best to describe the attributes! Images could be low resolution.
            DO NOT SHOW ANY UNNECESSARY MESSAGES!
            -------------------------------------------
            image list: {image_names[start:start + num]}"""
    },
]
\end{minted}
\caption{Prompt for getting CelebA predictions.}
\label{code:celeb}
\end{listing}

\begin{listing}[h]
\relsize{-3}
\centering
\begin{minted}{python}
content = [
    {
        "type": "text",
        "text": f"""Give the predictions of 
        [Clean_shaven, Chin_area, Side_to_side, Beard_area-Info_not_vis,
            Bread_length-5_o_clock_shadow, Bread_length-Short, Bread_length-Medium,
            Bread_length-Long, Bread_length-Info_not_vis, Mustache-None, Mustache-Isolated, Mustache-Connected_to_beard,
            Mustache-Info_not_vis, Sideburns-None, Sideburns_present, Sideburns-Connected_to_beard,
            Sideburns-Info_not_vis, Bald-FALSE, Bald-Top_only, Bald-Top_and_sides, Bald-Sides_only,
            Bald-Info_not_vis] for each image. Try your best to predict.
            to describe each image with True or False. "
           "You can just give the results (True/False) but should be in the correct order and use single space "
           "to separate each output."
            Output format: 'image name - No_Beard,Wavy_Hair,...,Bald' for each image.
            Self-check: 
            1) make sure you consider the logical relationship.
            Try you best to describe the attributes! Images could be low resolution.
            DO NOT SHOW ANY UNNECESSARY MESSAGES!
            -------------------------------------------
            image list: {image_names[start:start + num]}"""
    },
]
\end{minted}
\caption{Prompt for getting FH37K/41K predictions.}
\label{code:fh}
\end{listing}

\clearpage
{\small
\bibliographystyle{ieee_fullname}
\bibliography{main}
}